\documentclass[11pt]{article}

\usepackage[top=1in, bottom=1in, left=1.0in, right=1.0in]{geometry}

\usepackage{wrapfig}
\usepackage[round]{natbib}

\usepackage[utf8]{inputenc}
\usepackage{lmodern}
\usepackage[T1]{fontenc}
\usepackage{url}
\usepackage{booktabs}
\usepackage{amsfonts}
\usepackage{nicefrac}
\usepackage{microtype}
\usepackage[table]{xcolor}
\definecolor{matchgray}{gray}{0.92}

\usepackage[disable,textsize=tiny]{todonotes}

\usepackage{graphicx}
\usepackage{subcaption}
\usepackage{threeparttable}
\usepackage{multirow}
\usepackage{float}
\usepackage{adjustbox}
\usepackage{pifont}
\usepackage{makecell}
\usepackage{enumitem}

\usepackage{amsmath}
\usepackage{amssymb}
\usepackage{mathtools}
\usepackage{amsthm}

\usepackage{hyperref}
\hypersetup{
  colorlinks=true,
  linkcolor=[rgb]{0.10,0.25,0.55},
  citecolor=[rgb]{0.10,0.40,0.25},
  urlcolor=[rgb]{0.45,0.10,0.35},
  pdftitle={DLM-One: Diffusion Language Models for One-Step Sequence Generation},
  pdfauthor={Tianqi Chen, Shujian Zhang, Mingyuan Zhou},
  pdfkeywords={diffusion language models, score distillation, one-step generation, efficient inference}
}
\usepackage[capitalize,noabbrev]{cleveref}

\usepackage{algorithm}
\usepackage{algorithmic}

\usepackage{titlesec}
\titlespacing*{\paragraph}{0pt}{0.5ex plus 0.2ex minus 0.1ex}{0.3em}
\theoremstyle{plain}

\theoremstyle{definition}

\theoremstyle{remark}

\newcommand{\ours}{DLM-One}

\newcommand{\sid}{(SiD;~\citealp{zhou2024score}) }
\newcommand{\tmin}{t_{\text{min}}}
\newcommand{\tmax}{t_{\text{max}}}
\newcommand{\tinit}{t_{\text{init}}}

\def\eps{{\epsilon}}

\newcommand{\cN}{\mathcal{N}}

\newcommand{\cL}{\mathcal{L}}
\newcommand{\cD}{\mathcal{D}}

\newcommand{\bbE}{\mathbb{E}}

\newcommand{\whitenoise}{\cN(\mathbf{0},\mathbf{I})}
\newcommand{\sgdsm}[1]{{#1}^{\text{sg}}_{\text{dsm}}}
\newcommand{\sgadv}[1]{{#1}^{\text{sg}}_{\text{adv}}}
\newcommand{\gsid}[1]{{#1}^{\text{g}}_{\text{sd}}}
\newcommand{\gadv}[1]{{#1}^{\text{g}}_{\text{adv}}}

\newcommand{\sz}[1]{}

\title{DLM-One:
Diffusion Language Models\\ for One-Step Sequence Generation}

\author{%
  Tianqi Chen, Shujian Zhang, Mingyuan Zhou
  \thanks{The University of Texas at Austin, Austin, TX 78712, USA}
    \\
  \texttt{\{tqch,szhang19\}@utexas.edu, mingyuan.zhou@mccombs.utexas.edu} \\
}

\date{}

\begin{document}

\maketitle

\begin{abstract}
This paper introduces \textit{\ours{}}, a score-distillation-based framework for one-step sequence generation with continuous diffusion language models (DLMs). \ours{} eliminates iterative refinement by aligning the scores of a student model's outputs with the score function of a pretrained teacher DLM in the forward-diffused noisy space. We demonstrate that our framework is architecture-agnostic and robust across diverse continuous manifolds, including standard token embedding spaces and logit simplex spaces. Through experiments on multiple representative DLMs, we show that \ours{} achieves up to $\mathord{\sim}2000\times$ speedup in sampling steps and $\mathord{\sim}500\times$ in wall-clock time, while maintaining competitive performance on benchmark text generation tasks. We further analyze failure modes in language-domain diffusion distillation and propose an adversarially-regularized two-stage training scheme to prevent student degeneration. Our findings position one-step score distillation as a viable path for the efficient deployment of continuous diffusion models operating in continuous space for natural language processing.
\end{abstract}

\section{Introduction}
\label{sec:intro}

Recent progress in large language models (LLMs) has been primarily driven by autoregressive (AR) modeling, where sequences are generated token by token in a left-to-right fashion~\citep{vaswani2017attention,radford2018improving,brown2020language,achiam2023gpt,chowdhery2022palm,team2023gemini,touvron2023llama,Bai2023QwenTR,grattafiori2024llama}. While AR models have demonstrated remarkable performance across a wide range of natural language processing (NLP) tasks, they suffer from several well-known limitations: exposure bias, error accumulation, lack of bidirectional context during generation, limited controllability in non-left-to-right scenarios, and inability to revise previously generated text %
 \citep{keskar2019ctrl,Dathathri2020Plug,li2022diffusion,
reid2022diffuser,kaddour2023challenges,zhang2023planner,bachmann2024pitfalls,berglund2024the}. Moreover, certain data distributions may be inherently challenging to capture with AR models but can be modeled more effectively by alternative non-AR approaches, such as energy-based models~\citep{lin2021limitations}. The sequential nature of token generation also imposes a fundamental bottleneck on inference speed, motivating the development of various acceleration techniques to reduce computational overhead~\citep{khoshnoodi2024comprehensive}. These limitations have spurred growing interest in non-AR paradigms—particularly diffusion language models (DLMs)—which offer a fundamentally different approach by enabling parallel decoding of entire sequences instead of generating them one token at a time.

In contrast to AR LMs, which rely on causal attention and require one function evaluation (NFE) per token, DLMs often apply bidirectional attention and can generate sequences of predefined length in parallel~\citep{li2022diffusion,strudel2022self,dieleman2022continuous,gong2023diffuseq}. Existing DLMs perform generation via iterative refinement, enabling all tokens in a sequence to interact with each other and allowing for holistic reasoning over the full sequence. The per-token computational cost of DLMs depends on both the NFEs used during the iterative refinement process and the length of the target sequence. By adjusting the sequence length during pretraining and the number of NFEs at inference time, DLMs offer flexible configurations to trade off generation quality and speed~\citep{li2022diffusion,he2023diffusionbert,li2023diffusion,lin2023text,zheng2024a,gao2024empowering}.

However, despite this theoretical advantage and a few promising results by closed-source DLMs, such as Mercury~\citep{labs2025mercury} and Gemini~\citep{google2026geminidiffusion}, it remains a significant challenge for open-source DLMs to either generate faster while matching the performance of AR models, or achieve better performance at a comparable model size~\citep{gulrajani2023likelihood,han2023ssd,mahabadi2024tess,gong2024scaling,gong2025diffucoder,nie2025scaling,nie2025large,wang2025diffusion}. Nevertheless, there is substantial potential to accelerate DLMs by significantly reducing the number of required NFEs—without sacrificing performance—through diffusion distillation techniques. Such techniques have recently shown notable success in speeding up continuous diffusion models for vision tasks~\citep{sauer2024adversarial,yin2024improved,zhou2024long}.

DLMs can be broadly categorized into two types: \textbf{discrete} and \textbf{continuous}. Discrete DLMs operate directly on categorical token spaces~\citep{hoogeboom2021argmax,austin2021structured,he2023diffusionbert,lou2024discrete}, aligning naturally with the symbolic nature of language. These models have demonstrated promising performance, \textit{e.g.}, on unconditional text generation tasks. However, they still suffer from prohibitively slow sampling—often requiring hundreds to thousands of steps—due to the lack of effective acceleration techniques tailored to discrete diffusion. In contrast, this issue is less prominent in the vision domain, where continuous diffusion models and corresponding acceleration methods~predominate.

Unlike discrete diffusion, continuous DLMs model the diffusion process in the embedding space, treating token representations as continuous vectors~\citep{li2022diffusion,yuan2022seqdiffuseq,gong2023diffuseq,ye2023dinoiser,gulrajani2023likelihood,gao2024empowering}. Their sampling process naturally supports controllability via auxiliary guidance~\citep{dhariwal2021diffusion,ho2022classifier}, and can be further accelerated while maintaining competitive performance~\citep{song2021denoising,lu2022dpmsolver,salimans2022progressive}. These properties make DLMs particularly appealing for real-world applications. Although they are arguably less aligned with the inherently discrete nature of language—which may explain their relatively limited adoption compared to discrete DLMs—they offer a key advantage: compatibility with a wide range of acceleration strategies developed in the vision domain, such as consistency distillation~\citep{song2023consistency,song2023improved,geng2024consistency} and score distillation~\citep{poole2023dreamfusion,wang2023prolificdreamer,luo2023diffinstruct,yin2023one,zhou2024score}. These methods enable one- or few-step generation with minimal quality degradation and, when enhanced with real data during distillation, can even surpass the teacher model~\citep{zhou2025adversarial}.

This prompts a key question: \textbf{\textit{Can similar substantial gains in sampling efficiency be realized in language generation?}} More specifically, can we generate a sequence of, \textit{e.g.}, 100 tokens through a single forward pass of the diffusion backbone network? This would correspond to 100 NFEs for AR LMs, and potentially even more for existing DLMs, where the exact count depends on the number of iterative refinement steps but often reaches into the hundreds.

If so, it opens a promising research direction: how to pretrain stronger continuous DLMs that are naturally amenable to distillation. Potential approaches include improving the word embedding space or jointly optimizing it during pretraining. In this work, we focus on distilling existing continuous DLMs pretrained in the word embedding space, using publicly available checkpoints or open-source implementations, while leaving the design and pretraining of improved, larger models for future exploration. Specifically, we choose continuous DLMs pretrained with DiffuSeq~\citep{gong2023diffuseq} and Plaid~\citep{gulrajani2023likelihood} as our primary teacher models. %

\begin{figure*}[t]
  \centering
  \includegraphics[width=0.9\textwidth]{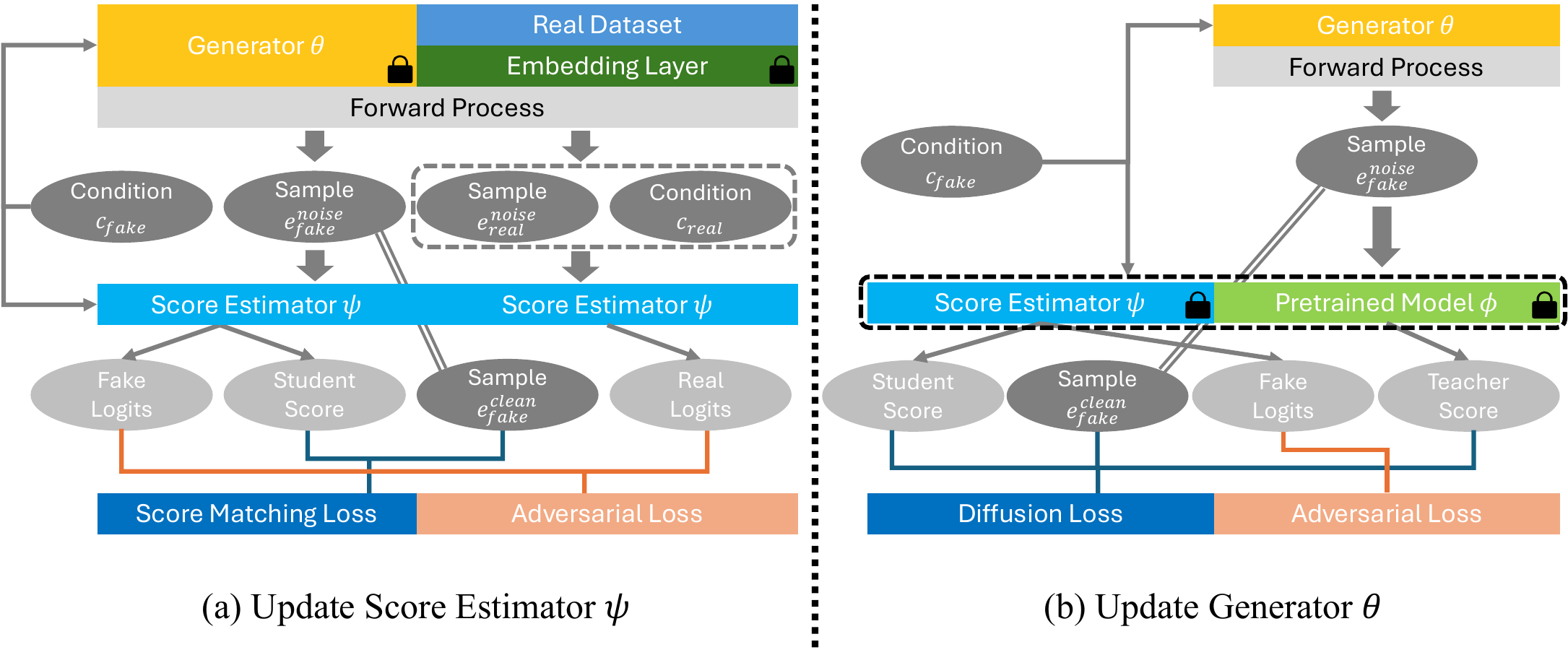}
  \vspace{-3mm}
  \caption{\small 
\textbf{Overview of the adversarial score distillation process.} 
\textbf{Left:} During score estimator $\psi$ updates, both real and generated data-condition pairs are used. 
The generator $\theta$ produces $e^{\text{clean}}_{\text{fake}}$ from $c_{\text{fake}}$, while real pairs are sampled from the dataset. 
The shared score estimator $\psi$ is trained for both score prediction and GAN discrimination. 
\textbf{Right:} During generator $\theta$ updates, the pretrained teacher model $\phi$ provides target scores, and $\psi$ produces both student scores and fake logits. These two scores are used to compute the score matching loss together with the clean data. 
Additionally, the generator is optimized to encourage the generation of more realistic samples under the feedback (i.e., logits) from $\psi$, via the adversarial loss. 
Modules marked with a \protect\includegraphics[height=1.8ex]{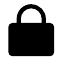} are frozen during the respective updates.
}\vspace{-4mm}
  \label{fig:overview}
\end{figure*}

We consider continuous diffusion for language modeling and investigate whether vision-inspired distillation techniques can enable drastically more efficient, high-quality sequence generation. Specifically, we propose a score distillation-based framework for training \textit{DLMs for one-step sequence generation} (\ours{}). Our method distills the knowledge of a pretrained teacher DLM into a student model of the same size that generates sequences in a single forward pass. Unlike prior work that often relies on hundreds of iterative refinement steps to produce a single sequence, \ours{} eliminates the need for iterative sampling altogether. It does so by aligning the scores of the student's outputs with the teacher’s score function in the forward-diffused noisy space. To stabilize training and prevent degenerate solutions, we introduce an auxiliary adversarial loss and adopt a two-stage optimization scheme that progressively refines the student.

Under the same model size, \ours{} achieves up to \( L \times \) speedup compared to AR LMs, where \( L \) is the target sequence length. It also achieves up to \( \mathrm{NFEs} \times \) speedup over the teacher DLM, where \( \mathrm{NFEs} \) denotes the number of iterative refinement steps used during teacher sampling. For example, in terms of wall-clock time, \ours{} delivers approximately \(500\times\) speedup over DiffuSeq, %
while achieving comparable generation quality. These results redefine what is possible along the Pareto front between generation quality and sampling efficiency.

Our contributions are summarized as follows:
\begin{itemize} %
[topsep=1ex,itemsep=0.5ex, parsep=0.5ex]
    \item We introduce \textbf{DLM-One}, a practical score distillation framework for continuous DLMs that enables one-step sequence generation without iterative denoising.
    \item We propose a two-stage training strategy with adversarial stabilization to enhance student quality and address common failure modes in distilling DLMs in a data-free setting.
    \item Our empirical evaluation across architectures and continuous spaces demonstrates that \ours{} achieves competitive performance while reducing sampling steps by up to $\mathord{\sim}2000\times$ and wall-clock time by up to $\mathord{\sim}500\times$ compared to standard iterative DLM teachers.

\end{itemize}

\section{Related Work}
\label{sec:related_work}

\subsection{Diffusion Language Models}
Unlike AR LMs, DLMs typically use a denoising score matching loss for training and predict entire sequences or multiple tokens at once. This eliminates the need for left-to-right, token-by-token sampling and enables faster decoding. Inspired by continuous diffusion models~\citep{ho2020denoising,nichol2021improved}, \citet{li2022diffusion} propose an end-to-end language modeling approach that jointly learns word embeddings and a diffusion model in the embedding space, combining a diffusion loss with a rounding loss. \citet{gong2023diffuseq} adopt a similar strategy for sequence-to-sequence tasks by concatenating conditioning inputs with target sequences and modifying the forward diffusion process to apply noise only to the target. In contrast to the decoder-only architecture used in DiffuSeq, \citet{yuan2022seqdiffuseq} introduce a dedicated encoder to process the conditioning input.

Viewing the additional rounding loss as a regularization term, \citet{gao2024empowering} propose an anchor loss to improve training stability and prevent embedding collapse. To bridge the likelihood gap, \citet{gulrajani2023likelihood} introduce Plaid, the first DLM shown to achieve likelihood performance comparable to that of AR models on standard language modeling benchmarks.
While these models all 
 operate in the embedding space, we note that DLMs have also been trained in the vocabulary logit space~\citep{han2023ssd,mahabadi2024tess} and the latent space of an encoder-decoder LM~\citep{lovelace2023latent,zhang2023planner,zhou2024diffusion,shabalin2025tencdm}. Extending DLM-One to such models represents a promising direction for future work.

In addition to continuous diffusion models, discrete diffusion models have also been studied for text generation. \citet{hoogeboom2021argmax} introduce a multinomial diffusion process for modeling categorical data. \citet{austin2021structured} further explore various discrete state transition matrices, adding flexibility to the discrete diffusion process. By vector quantizing images into sequences of visual tokens~\citep{oord2017neural,esser2021taming}, discrete diffusion models have been applied to generate visual token sequences that can be decoded back into images~\citep{gu2021vector,hu2022global}.
 \citet{lou2024discrete} extend score matching \citep{hyvarinen2005estimation} losses from continuous to discrete spaces. \citet{ou2025your} reformulate the concrete score \citep{meng2022concrete} as a product of time-independent conditional probabilities and a time-dependent scalar, enabling more efficient sampling. Rather than working on the general forward process, \citet{sahoo2024simple} improve the practical performance of discrete DLMs by focusing on the masking strategy and introducing tight Rao-Blackwellized objectives. \citet{shi2024simplified} derive a simplified variational objective for continuous-time masked DLMs and generalize the masking schedule to support state dependency. 
Recognizing the connection between masked DLMs and AR models, \citet{gong2024scaling} propose a continual pretraining approach to adapt pretrained AR models into discrete DLMs. \citet{nie2025large} introduce LLaDA that pretrains a large discrete DLM from scratch and further improves it with supervised fine-tuning.

\subsection{Faster Diffusion} 
Diffusion models are known for their strong generative capabilities; however, this comes at the cost of hundreds to thousands of NFEs during sampling in their original formulation~\citep{ho2020denoising, song2020score}. Despite progress with training-free acceleration techniques, such as advanced samplers~\citep{liu2022pseudo, lu2022dpmsolver} and model quantization~\citep{li2023q}, diffusion models still lag behind traditional generative models like GANs and VAEs in terms of sampling speed.

Several directions have been explored to accelerate diffusion-based generation. \citet{liu2024discrete} and \citet{guo2024plug} propose Discrete Copula Diffusion, which combines a discrete diffusion model with a copula-based correction module at inference time to improve the denoising distribution. Masked diffusion models (MDMs)~\citep{zheng2024masked} accelerate generation via a first-hitting sampling strategy. Progressive distillation~\citep{salimans2022progressive} introduces an iterative distillation scheme, reducing the number of sampling steps by progressively halving them. \citet{luo2023diffinstruct} and \citet{yin2024improved} propose minimizing the integral Kullback–Leibler divergence between the generative distributions of teacher and student models. From a score-distillation perspective, \citet{zhou2024score} propose a Fisher divergence-based distillation objective and an accompanying alternating optimization procedure that jointly enhance convergence and generation quality. Further improvements in data-free score distillation have been achieved by incorporating real data and adversarial training~\citep{sauer2024adversarial, yin2024improved, zhou2025adversarial}.

In the context of accelerating DLMs, AR-Diffusion~\citep{wu2023ar} incorporates autoregressive characteristics into diffusion models by allocating fewer refinement steps to earlier tokens, thereby better modeling sequential dependencies. Unlike training-free methods that focus on better utilizing the frozen teacher for faster inference, %
diffusion distillation trains a student model from a pretrained teacher, enabling generation in just one or a few inference steps. 
Our work—\textit{DLM-One}—is a diffusion distillation framework that enables one-step sequence generation while preserving the generation quality of the teacher, effectively eliminating the need for iterative refinement.

\section{One-Step Diffusion Language Models}
\label{sec:osdlm}

To train a one-step sequence generation model, we begin with a pretrained teacher DLM that operates in a continuous embedding space. In this setting, each discrete language token is first mapped to a real-valued embedding vector via an embedding layer. The diffusion process is then applied to these continuous embeddings rather than to the discrete tokens themselves. This setup enables us to leverage well-established acceleration methods from continuous diffusion models in the vision domain, while focusing on language-model-specific adjustments essential for effective sequence~generation.

During pretraining, the embedding matrix is typically optimized end-to-end to improve generation quality~\citep{li2022diffusion}, as this allows the embeddings to better align with the denoising objective compared to using a frozen embedding matrix from a pretrained language model. However, without additional constraints, the embedding space can exhibit pathological behaviors such as collapse or poor token separation. To address this, recent work has proposed regularization techniques—such as anchor loss and likelihood-aware training—to preserve meaningful structure in the embedding space~\citep{gong2023diffuseq,gulrajani2023likelihood,gao2024empowering}.

\subsection{Embedding-space Score Distillation}

Following the practice adopted in latent diffusion~\citep{rombach2022high}, we freeze the pretrained embedding matrix during distillation, leaving end-to-end embedding finetuning as a promising direction for future work. While various objectives are possible, we build our method upon Score identity Distillation~\sid to demonstrate the potential of one-step diffusion models in the language domain. SiD is a state-of-the-art one-step diffusion distillation method that operates in a fully data-free setting and readily supports two key enhancement techniques—classifier-free guidance (CFG)~\citep{zhou2024long} and adversarial training~\citep{zhou2025adversarial}—both of which are found to be important for distillation in the embedding space of DLMs.

Specifically, we denote the pretrained teacher DLM as $\phi$, the student generator as $\theta$, and the score estimator for the student model as $\psi$. Let $E$ denote the token embedding layer and $e \in \mathbb{R}^{d \times L}$ denote the $d$-dimensional continuous embeddings of a sequence of length $L$, which may optionally be mapped back to discrete tokens via a rounding or decoding mechanism during inference. The generation process of the student model is given by
\[
e = G_\theta(c, z), \quad z \sim \mathcal{N}(0, \mathbf{I}),
\]
where $c$ is an optional condition ($e.g.$, a prompt or label), and $z$ is noise input. %
We apply the forward diffusion process to obtain noisy embeddings
\(
e_t = \alpha_t e + \sigma_t \epsilon, ~ \epsilon \sim \mathcal{N}(0, \mathbf{I}),
\)
where $\alpha_t$ and $\sigma_t$ follow a predefined noise schedule that gradually decreases the signal-to-noise ratio $\alpha_t / \sigma_t$ as $t$~increases.

The pretrained teacher model $\phi$ provides an estimate of the score function at $e_t$ given $t$ and $c$, defined as
$
s_\phi(e_t, t, c) = \nabla_{e_t} \log p(e_t \mid t, c).
$
The distillation objective is to train the student generator such that its score matches that of the teacher in the forward-diffused noisy space. This is achieved by minimizing the model-based explicit score matching (MESM) loss, a form of Fisher divergence:
\begin{align}
   \mathcal{L}_{\text{mesm}}(\theta) = \mathbb{E}_{e,t,c} \left[ \omega_t \left\| s_\phi(e_t, t, c) - s_{\psi^*(\theta)}(e_t, t, c) \right\|^2 \right],
   \label{eq:mesm}
\end{align}
where $\psi^*(\theta)$ denotes the true score function induced by the student generator $\theta$, and $\omega_t$ is a time-dependent reweighting coefficient. For unconditional generation, the condition $c$ is set to $\emptyset$.

By Tweedie's formula~\citep{robbins1992empirical,efron2011tweedie}, Equation~\ref{eq:mesm} can be equivalently written as:
\begin{align}
 \textstyle   \mathbb{E}_{e,t,c} \left( \omega_t \frac{\alpha_t^2}{\sigma_t^4} \| \hat{e}_\phi(e_t, t, c) - \hat{e}_{\psi^*(\theta)}(e_t, t, c) \|^2 \right),
   \label{eq:mdsm-tweedie}
\end{align}
where $\hat{e}_\phi$ and $\hat{e}_{\psi^*(\theta)}$ denote the expected values of the clean embedding $e$ conditioned on the noisy observation $e_t$, as inferred by the teacher and optimal student score networks, respectively.

While Equation~\ref{eq:mdsm-tweedie} and its gradient are generally intractable, the SiD method \citep{zhou2024score} provides an effective optimization procedure that alternates between estimating $\psi^*(\theta)$ and updating $\theta$. Specifically, we optimize $\psi$ given $\theta$ using the denoising score matching (DSM)~loss:
\begin{align}
   \mathcal{L}_{\text{dsm}}(\psi) = \mathbb{E}_{e, t, c} \left[ \gamma_t \left\| \hat{e}_\psi(e_t, t, c) - e \right\|^2 \right],
   \label{eq:dsm}
\end{align}
and optimize $\theta$ given $\psi$ using the following SiD loss $\mathcal{L}_{\text{sid}}(\theta; \psi^*, \mu)$, defined as:
\begin{align}
&\textstyle \mathbb{E}_{e, t, c} \big[ (1 - \mu) \, \omega_t \frac{\alpha_t^2}{\sigma_t^4} \left\| \hat{e}_\phi(e_t, t, c) - \hat{e}_\psi(e_t, t, c) \right\|^2 \notag \\
&\textstyle  \quad + \omega_t \frac{\alpha_t^2}{\sigma_t^4} \left( \hat{e}_\phi(e_t, t, c) - \hat{e}_\psi(e_t, t, c) \right)^\top \left( \hat{e}_\psi(e_t, t, c) - e \right) \big],
   \label{eq:sid}
\end{align}
where $\mu>0$ is a hyperparameter that %
is often set as 1 or 1.2.

\subsection{Adversarial Regularization}
\label{sec:advreg}

While data-free distillation of pretrained diffusion models is appealing—requiring access only to the teacher model rather than real data—and has achieved highly competitive performance in the vision domain \citep{zhou2024score,zhou2024long}, its application to DLMs presents a major challenge: degeneration in the student model. In the absence of explicit constraints ($e.g.$, on sentence length) or implicit supervision from real data, distilled models tend to degenerate after a certain number of training iterations, such as (1) generating repetitive tokens, or (2) producing empty sequences filled with \texttt{[PAD]} tokens. To mitigate this, we combine standard score distillation with adversarial regularization.

Specifically, when updating the fake score estimator $\psi$, we first sample a condition $c^{\text{fake}}$ and generate an embedding sequence $e_\theta^{\text{fake}}$ using the student generator $\theta$. We then compute the DSM loss of $\psi$ along with part of the adversarial loss—namely, the binary cross-entropy (BCE) loss using pseudo-labels set to all negatives. Additionally, we sample a pair consisting of a real data sequence $x^{\text{real}}$ and its corresponding condition $c^{\text{real}}$, and compute the remaining part of the adversarial loss using pseudo-labels set to all positives. Following Diffusion GAN \citep{wang2022diffusion} to perform discrimination on noised embeddings, the adversarial 
loss for $\psi$ is given by:
\begin{align}
    \textstyle \frac{1}{2}\bbE\big[\log\sigma( D_\psi(e_t^{\text{real}},t, c^{\text{real}}))+\log(1-\sigma( D_\psi(e_{\theta,t}^{\text{fake}},t,c^{\text{fake}})))\big],\label{eq:sg_adv}
\end{align}
where $e_t^{\text{real}}$ and $e_{\theta,t}^{\text{fake}}$ are the noisy embeddings obtained by forward diffusing the embedding of $x^{\text{real}}$ and $e_{\theta}^{\text{fake}}$, respectively. $D_{\psi}$ is some scalar function derived from $\psi$.
For the update steps of the student model $\theta$, we compute both the SiD loss and the all-positive BCE loss on generated sequences conditioned on $c$. We denote each generated \(\langle\text{data}, \text{condition}\rangle\) pair as \((x_\theta, c)\) and $e_{\theta,t}$ as the noised version of $e_\theta$. The corresponding adversarial loss is:
\begin{align}
    \cL_{\text{adv}}^{\text{g}}(\theta)=\bbE\left[\log \sigma(D_\psi(e_{\theta,t},t,c))\right].\label{eq:g_adv}
\end{align}

We provide an overview and pseudo-code of our adversarial score distillation training process in \Cref{fig:overview} and \Cref{algo:main}, respectively. For efficiency, we utilize the same model (\textit{i.e.,} the score estimator $\psi$) for both score prediction and GAN discrimination. At a high level, the additional adversarial losses provide implicit supervision and help stabilize training, preventing mode collapse and encouraging more realistic sequence generation.

\subsection{Two-stage Training}
\label{sec:twostage}

Due to the alternating update scheme, the score estimator $\psi$ may fail to accurately approximate the true score of the student model $\theta$. We address this by proposing a two-stage training procedure. In \textit{Stage 1}, we focus on obtaining a ``good enough'' student model whose generative distribution is reasonably close to the teacher’s. In practice, we train the student for a fixed number of steps and select the best checkpoint based on validation BLEU scores. In \textit{Stage 2}, we resume training $\theta$ from that checkpoint but reinitialize the score estimator $\psi$ with the teacher model's parameters $\phi$. This mitigates the potential lag of $\psi$, which occurs when the estimator falls behind the true score of the evolving student. This lag is more pronounced as the student’s distribution approaches the teacher’s and diverges from its initial state, causing the feedback from $\psi$ to become insufficient. Reinitializing $\psi$ realigns it with the updated student, providing more meaningful learning signals. This stage mirrors \Cref{algo:main} but requires the Stage 1 checkpoint for initialization.

\section{Experiments}
\label{sec:experiments}

In our experiments, we conduct a comprehensive evaluation on the benchmark tasks originally used to assess the performance of the teacher DLMs pretrained with DiffuSeq and Plaid.
The results convincingly demonstrate the potential of significantly accelerating the sampling efficiency of continuous DLMs via score distillation, enabling one-step token sequence generation that rivals the performance of teacher models requiring hundreds of times more computation. This redefines the Pareto frontier between computational efficiency and generation quality in continuous diffusion-based language modeling, and has profound implications for the future development of LLMs.

\begin{table*}[t]
\centering
\caption{\small \textbf{Performance comparison between teacher and student models across Seq2Seq tasks.} $\uparrow$ indicates higher is better, $\downarrow$ indicates lower is better. 
\textsuperscript{*} denotes that the student's performance is within 5\% of the teacher's, and 
\textsuperscript{**} indicates that it is within 1\%.
}
\vspace{-1mm}
\label{tab:main_results}
\footnotesize
\begin{adjustbox}{max width=\linewidth}
\begin{tabular}{llllllcc}
\toprule
\textbf{Task} & \textbf{Model} & \textbf{BLEU}($\uparrow$) & \textbf{ROUGE-L}($\uparrow$) & \textbf{BERT}($\uparrow$) & \textbf{Dist-1}($\uparrow$) & \textbf{SelfBLEU}($\downarrow$)~/~\textbf{Div-4}($\uparrow$) & \textbf{NFEs}($\downarrow$) \\
\midrule

\multirow{2}{*}{PP} 
& DiffuSeq
& 0.1829 
& 0.5299
& 0.7932
& 0.9747
& 0.2732~/~0.8641
& 2000 \\
& \ours{}
& $0.1788^{*}$
& $0.5265^{**}$
& $0.7851^*$
& $0.9671^{**}$
& 0.3418~/~0.6256
& \textbf{1} \\

\midrule

\multirow{2}{*}{QG} 
& DiffuSeq 
& 0.1512
& 0.3468
& 0.5871
& 0.9141
& 0.2789~/~0.8103
& 2000 \\
& \ours{} 
& $0.1512^{**}$
& 0.3257
& $0.5683^*$
& $0.9053^{**}$
& 0.6166~/~0.3798
& \textbf{1} \\

\midrule

\multirow{2}{*}{TS} 
& DiffuSeq 
& 0.2929
& 0.5313 
& 0.7781
& 0.9272 
& 0.4642~/~0.6604
& 2000 \\
& \ours{}
& $0.2927^{**}$
& $0.5299^{**}$
& $0.7565^*$
& $0.8924^*$
& 0.5456~/~0.4098
& \textbf{1} \\

\bottomrule
\end{tabular}
\end{adjustbox}
\vspace{-2mm}
\end{table*}

\subsection{Tasks and Datasets}

 We consider three sequence-to-sequence~(Seq2Seq) tasks, including: question generation (QG), text simplification (TS), and paraphrase (PP). Specifically, we used preprocessed data from $\text{Quasar-T}$~\citep{dhingra2017quasar} for QG, Wiki-Auto~\citep{jiang2020neural} for TS, and Quora question pairs~(\href{https://www.kaggle.com/c/quora-question-pairs}{QQP}) for PP.
For each dataset, we use the standard splits of training, validation, and test sets. The data derived from Quasar-T contain approximately 129k $\langle$document, question$\rangle$ pairs, including 117k training pairs, 2k validation pairs, and 10k test pairs. The Wiki-Auto preprocessed dataset consists of a total of $\sim$685k $\langle$complex, simple$\rangle$ sentence pairs, with approximately 678k training pairs, 2k validation pairs, and 5k test pairs. The QQP dataset contains about 150k paraphrase sentence pairs, including 145k training, 2k validation, and 2k test. 

In addition to Seq2Seq tasks, we also conduct experiments on unconditional text generation. For this setting, we follow the setup introduced by \citet{gulrajani2023likelihood} and use the OpenWebText2~\citep{gao2020pile} dataset, which consists of high-quality English web content filtered to resemble the pretraining corpus of GPT-2. The dataset contains approximately 10 million documents, and we use a 1B-token subset for training, consistent with prior work.

\subsection{Evaluation} 
For evaluation of the Seq2Seq tasks, we mainly consider five factors: BLEU~\citep{papineni2002bleu}, ROUGE-L~\citep{lin-2004-rouge}, BERT Score~\citep{Zhang*2020BERTScore:}, Dist-1, and sequence diversity. BLEU, ROUGE-L, and BERTScore are standard metrics for evaluating sequence-to-sequence tasks, as they capture sentence-level similarity between the generated sequences and the references. BLEU emphasizes n-gram precision, ROUGE-L focuses on recall based on the longest common subsequence, and BERTScore leverages contextual embeddings to assess semantic similarity. Dist-1 measures lexical diversity by computing the average ratio of distinct unigrams in a single sentence over all generated samples. Sequence-level diversity is further assessed using two metrics: self-BLEU \citep{zhu2018texygen} and Div-4. Following the implementation of DiffuSeq~\citep{gong2023diffuseq}, we compute self-BLEU by averaging inter-sentence BLEU scores across generated samples, while Div-4 quantifies the proportion of distinct 4-grams among them. For the unconditional generation task, we assess the model performance mainly through the generative perplexity. Specifically, we generate 250 samples and calculate the average perplexity evaluated under GPT-2~\citep{radford2019language}.

\begin{figure*}[!htbp]
  \centering
  \begin{subfigure}[b]{0.225\textwidth}
    \centering
    \includegraphics[width=\linewidth]{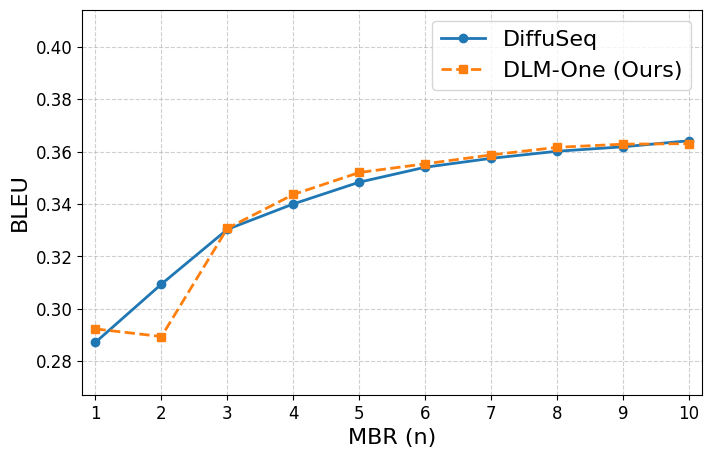}
    \caption{BLEU}
  \end{subfigure}
  \begin{subfigure}[b]{0.225\textwidth}
    \centering
    \includegraphics[width=\linewidth]{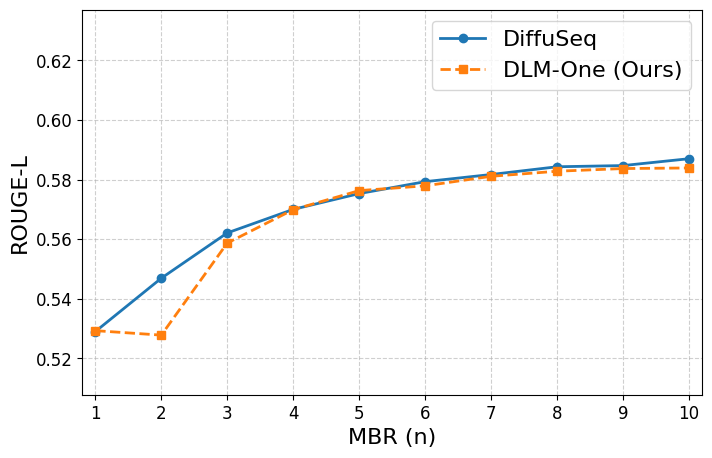}
    \caption{ROUGE-L}
  \end{subfigure}
  \begin{subfigure}[b]{0.225\textwidth}
    \centering
    \includegraphics[width=\linewidth]{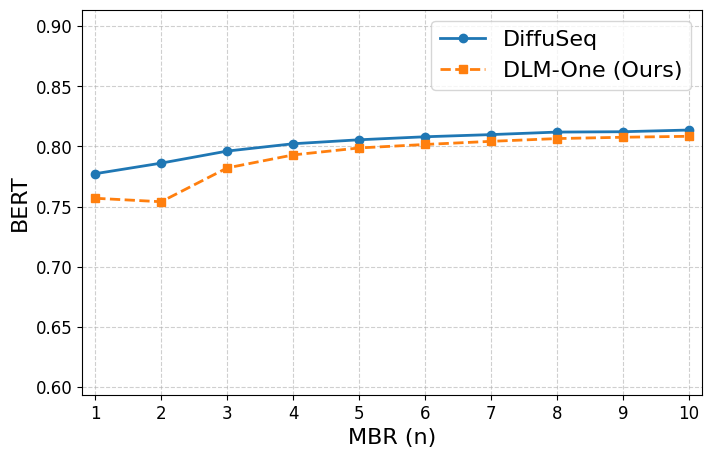}
    \caption{BERTScore}
  \end{subfigure}
  \begin{subfigure}[b]{0.225\textwidth}
    \centering
    \includegraphics[width=\linewidth]{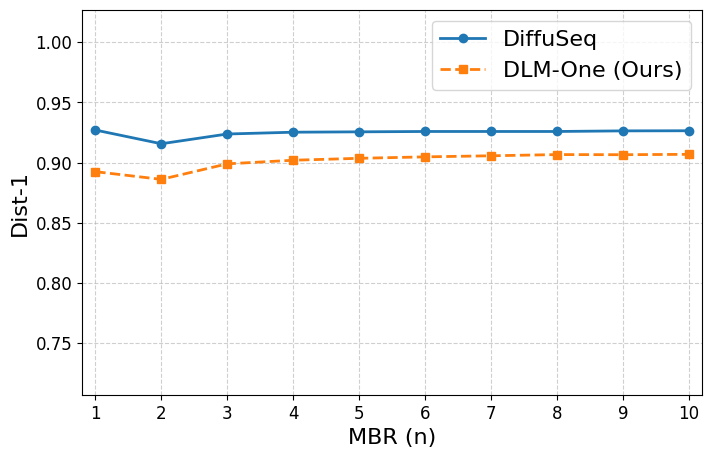}
    \caption{Dist-1}
  \end{subfigure}

  \caption{Evaluation metrics using MBR decoding across 1 to 10 candidate(s) on the Wiki dataset.}
  \label{fig:wiki_mbr_all}
  \vspace{-4mm}
\end{figure*}

\subsection{Sequence-to-Sequence~(Seq2Seq) Tasks}

\paragraph{Baselines and Generality.} 
For Seq2Seq tasks, we mainly consider \textbf{DiffuSeq}~\citep{gong2023diffuseq} as our major baseline to showcase the effectiveness of the proposed score distillation framework for LMs. Additionally, we investigate the general applicability of \ours{} by extending the framework to another continuous DLM architecture, \textbf{TESS}~\citep{mahabadi2024tess}, which operates on a logit simplex manifold rather than a traditional embedding space. We include the results of this case study in Appendix~\ref{app:tess_details}. We list results of all five performance metrics for our primary baselines in \Cref{tab:main_results}, which shows that our distilled models can achieve close-to-teacher performance consistently across all three tasks while taking a significantly smaller number of function evaluations (NFEs).

\begin{figure*}[t]
  \centering
  \begin{subfigure}[b]{0.3\textwidth}
\includegraphics[width=\linewidth]{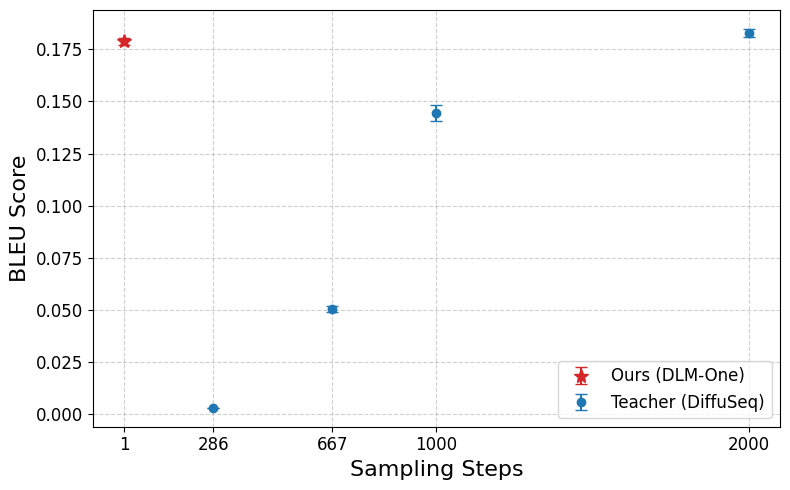}
    \caption{QQP}
\end{subfigure}
  \hfill
  \centering
  \begin{subfigure}[b]{0.3\textwidth}
  \includegraphics[width=\linewidth]{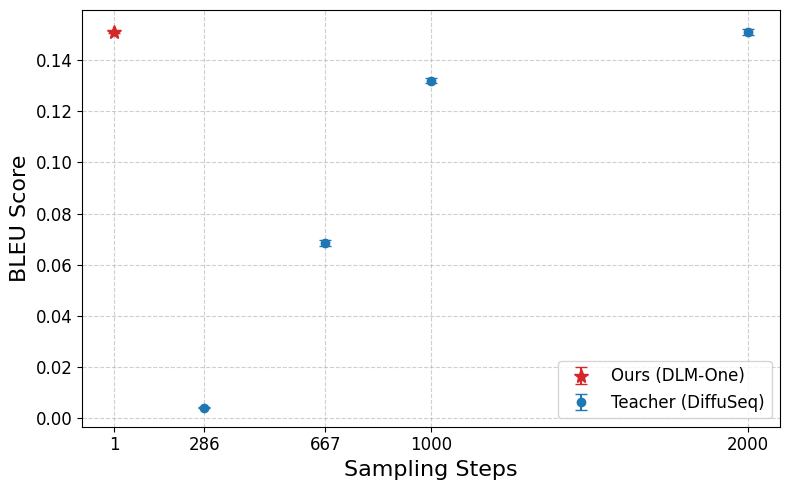}
    \caption{Quasar-T}
\end{subfigure}
  \hfill
\centering
  \begin{subfigure}[b]{0.3\textwidth}
  \includegraphics[width=\linewidth]{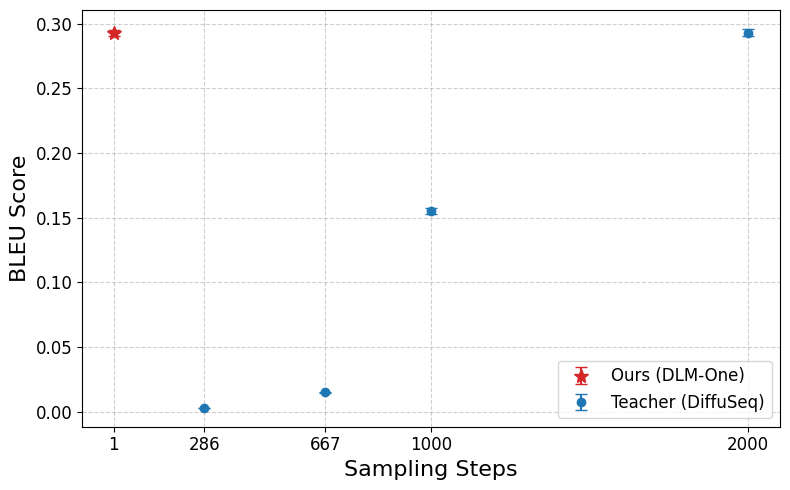}
    \caption{Wiki-Auto}
\end{subfigure}
  \caption{\small 
  \textbf{BLEU score vs. sampling steps on different datasets.} 
  The teacher model (DiffuSeq) requires hundreds to thousands of denoising steps to reach optimal performance, while our \ours{} achieves competitive BLEU in a single step—offering over \textbf{100× faster} generation without significant quality degradation.
  }
  \label{fig:bleu_vs_steps}
  \vspace{-2mm}
\end{figure*}

\paragraph{Inference Acceleration.} 
The actual acceleration is further demonstrated in \Cref{fig:bleu_vs_steps}, where we consider the BLEU score against the number of sampling steps on QQP, Quasar-T, and Wiki-Auto datasets. In \Cref{tab:inference_time}, we provide the conversion from sampling steps to inference time, measured on an NVIDIA RTX A5000 GPU. Our one-step model achieves up to an approximately \textbf{500$\times$} speedup in terms of wall-clock time compared to the 2000-step teacher baseline with no notable performance degradation.

\paragraph{Training Stages and MBR Decoding.}
The results of our approach on PP and QG are obtained from final-stage (i.e., \textit{Stage 2}) \ours{} models, while those on TS are reported from \textit{Stage 1}, as the student model already closely matches the teacher’s performance. As shown in \Cref{fig:wiki_mbr_all}, minimum Bayes risk (MBR) decoding offers a more comprehensive evaluation of generation quality and diversity by leveraging multiple candidate samples. As the number of candidates increases, MBR decoding typically leads to improved performance. The observation that our student consistently matches the teacher across 1 to 10 MBR decoding candidates further suggests that single-stage distillation is sufficient for the TS task on the Wiki dataset.

\subsection{Unconditional Text Generation}
To assess the generality and scalability of \ours{}, we further conduct experiments on a more complex text generation task and a larger-scale dataset. Specifically, we select \textbf{Plaid}~\citep{gulrajani2023likelihood} as our baseline for unconditional text generation on the OpenWebText2 dataset. We configure the teacher Plaid model to use a 16-block transformer backbone with 384 hidden dimensions and 6 attention heads. As shown in \Cref{tab:openweb_perplexity}, our \ours{} model achieves competitive performance compared to the multistep Plaid teacher model with up to 256 inference steps, which is effectively a 256$\times$ speedup.

\begin{table}[t]
    \centering
    \caption{\small 
    \textbf{Average generation time per sample across different sampling steps of DiffuSeq.} Each entry reflects the mean time (in seconds), averaged over 100 runs. Time does not scale strictly linearly with NFEs, due to fixed overhead such as embedding rounding and tokenizer-based decoding.}
    \label{tab:inference_time}
    \vspace{-1mm}
    \small
\begin{adjustbox}{max width=\linewidth}
    \begin{tabular}{lcccccc}
    \toprule
    \textbf{Steps} & 1 & 65 & 286 & 667 & 1000 & 2000 \\
    \midrule
    \textbf{Time (s)} & 0.03 & 0.51 & 2.25 & 5.20 & 7.70 & 14.94 \\
    \bottomrule
    \end{tabular}
    \end{adjustbox}
\end{table}

\begin{table}[t]
    \centering
    \caption{\small
    \textbf{Mean perplexity of the generated samples output by \ours{} (Student) and Plaid (Teacher) using different inference steps.} The results correspond to the unconditional text generation task evaluated under GPT-2.}
    \label{tab:openweb_perplexity}
    \vspace{-1mm}
    \small
\begin{adjustbox}{max width=\linewidth}
    \begin{tabular}{cccccc}
    \toprule
    \textbf{Model}
    & \multicolumn{1}{c}{\textbf{Student}} 
    & \multicolumn{4}{c}{\textbf{Teacher}} \\
    \cmidrule(lr){1-1} \cmidrule(lr){2-2} \cmidrule(lr){3-6}
    \textbf{\# Inf. Steps} & 1 & 16 & 64 & 256 & 4096 \\
    \textbf{Perplexity} & \underline{93.99} & 298.92 & 122.41 & \underline{94.28} & \textbf{83.37} \\
    \bottomrule
    \end{tabular}
    \end{adjustbox}
    \vspace{-4mm}
\end{table}

\begin{table}[t]
    \caption{\small \textbf{Effect of two-stage training on the QQP dataset.} 
    The second row shows raw scores; the third row shows relative changes from Stage 1.
    Percentages in {\color{green!70!black}green} and {\color{red!70!black}red} indicate improvements and degradations, respectively.
    Arrows $\uparrow$/$\downarrow$ denote preferred directions.}
    \vspace{-1mm}
    \label{tab:twostage_effect}
    \centering
    \small
\begin{adjustbox}{max width=\linewidth}
    \begin{tabular}{l|cccccc}
    \toprule
    \textbf{Stage} & \textbf{BLEU}($\uparrow$) & \textbf{ROUGE-L}($\uparrow$) & \textbf{BERT}($\uparrow$) & \textbf{Dist-1}($\uparrow$) & \textbf{SelfBLEU}($\downarrow$) & \textbf{Div-4}($\uparrow$) \\
    \midrule
    Stage 1        & 0.1468 & 0.4829 & 0.7402 & 0.9370 & 0.2195 & 0.7764 \\
    Stage 2        & 0.1788 & 0.5265 & 0.7851 & 0.9671 & 0.3418 & 0.6256 \\ \midrule
    $\Delta$Stage 
                  & {\color{green!70!black}+21.8\%} 
                  & {\color{green!70!black}+9.0\%} 
                  & {\color{green!70!black}+6.1\%} 
                  & {\color{green!70!black}+3.2\%} 
                  & {\color{red!70!black}+55.7\%} 
                  & {\color{red!70!black}$-$19.4\%} \\
    \bottomrule
    \end{tabular}
    \end{adjustbox}
    \vspace{-4mm}
\end{table}

\section{Discussion}
\paragraph{Analysis of Model Degeneration.} When directly applying data-free diffusion distillation methods for vision models to DLMs, we noticed that the student model will inevitably suffer from inferior generation quality and even degeneration. After thoroughly reviewing the failure patterns and inspecting the deeper causes, we identify two major issues specific to DLM distillation: 
\begin{enumerate}[nosep, leftmargin=*]
    \item \textbf{Poor Initial Predictions.} Unlike in the vision domain, where diffusion model's predictions at a large timestep can be blurry but still recognizable, a DLM’s initial predictions are often incoherent (\textit{e.g.,} gibberish and making little sense), exacerbating the inaccurate score estimation problem of the fake score network and restricting the student's final performance (see also \Cref{sec:twostage}). Therefore, in DLM-One, we propose a two-stage training approach to mitigate this initial score mismatch issue in the second stage.
    \item \textbf{Variable Output Length.} Unlike multistep models, one-step DLMs must determine the final output length upfront, which makes training unstable and prone to degeneration. One specific issue is related to the use of trailing \texttt{[PAD]} tokens. Without supervision from data, the student can easily fail by simply learning to output a target sequence full of \texttt{[PAD]} tokens, because it can trick the teacher model into treating it as a ``valid'' data pattern and lead to a very small discrepancy between the teacher model and the fake score model. This issue is also noted in \Cref{sec:advreg}, where we introduce an adversarial loss term to provide regularization on the generation sequence length.
\end{enumerate}

\paragraph{Effect of Two-stage Training.}\label{dis:twostage} 
We find the two-stage training strategy is crucial for improving the model's overall fidelity across key metrics such as BLEU, ROUGE-L, and BERTScore, at the cost of reduced diversity. For practical applications of DLM-One, we argue this is a favorable trade-off, as higher fidelity often corresponds to greater model utility for end-users. The trade-off is further quantified in \Cref{tab:twostage_effect}, which compares the performance of the QQP checkpoints from the two stages. 
Furthermore, we demonstrate that this loss in diversity can be mitigated with inference-time augmentation, and we also discuss the possibility of generalizing DLM-One to a few-step model, as detailed in Appendix~\ref{app:tradeoff}.

\begin{figure}[t]
  \centering
  \begin{subfigure}[b]{0.22\textwidth}
    \centering
\includegraphics[width=\linewidth]{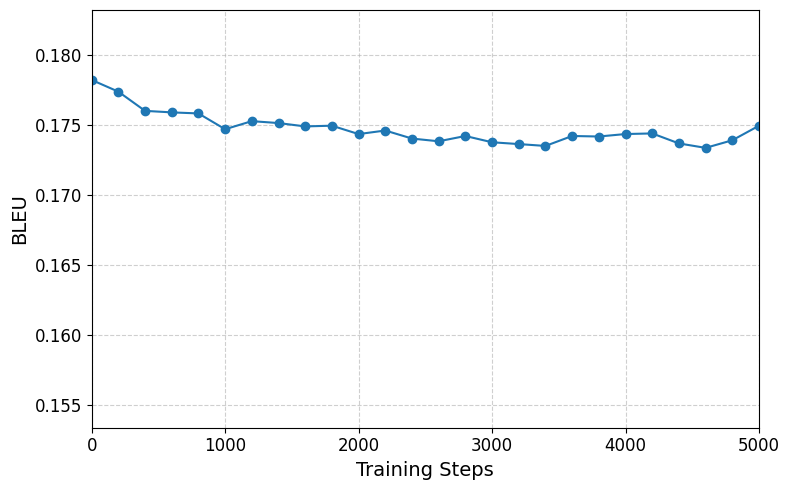}
    \caption{BLEU}
  \end{subfigure}
  \begin{subfigure}[b]{0.22\textwidth}
    \centering
    \includegraphics[width=\linewidth]{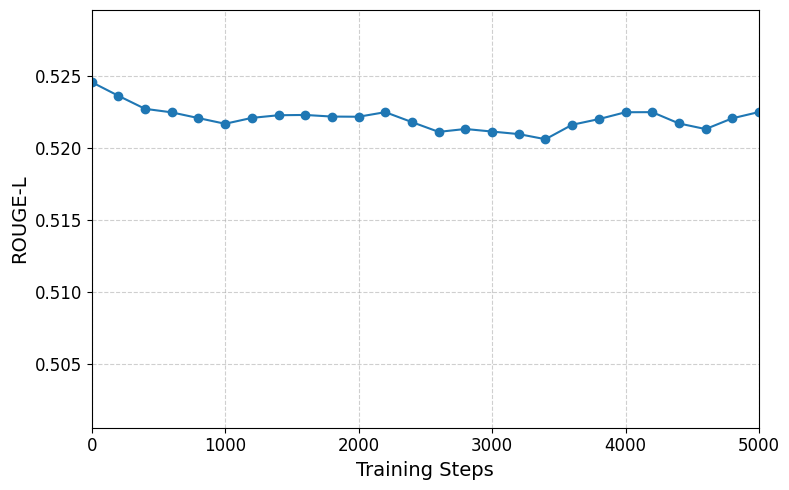}
    \caption{ROUGE-L}
  \end{subfigure}
  \begin{subfigure}[b]{0.22\textwidth}
    \centering
    \includegraphics[width=\linewidth]{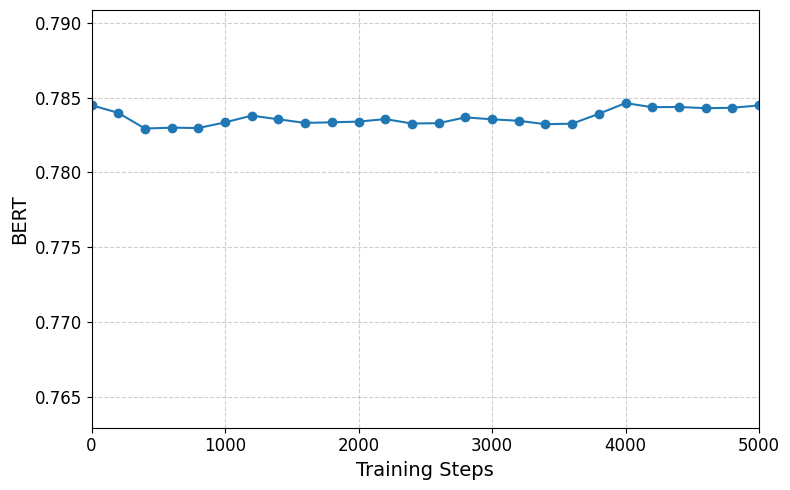}
    \caption{BERTScore}
  \end{subfigure}
  \begin{subfigure}[b]{0.22\textwidth}
    \centering
    \includegraphics[width=\linewidth]{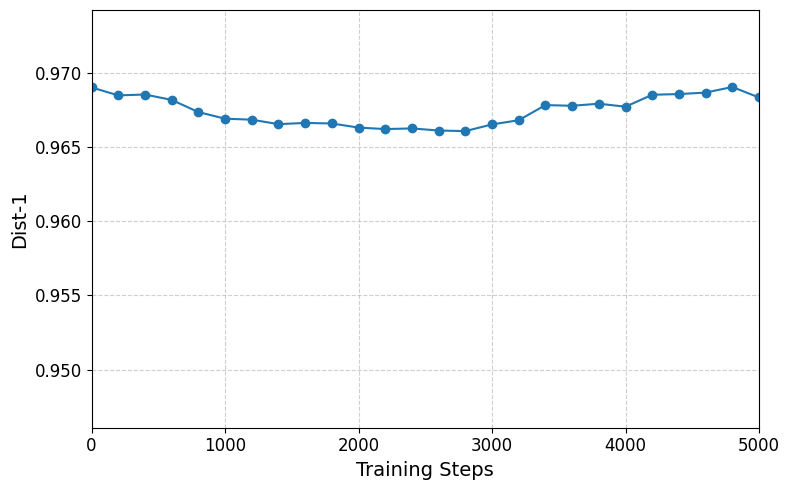}
    \caption{Dist-1}
  \end{subfigure}
  \caption{\small 
  Evolution of evaluation metrics during Stage 3 training on the QQP dataset.}
  \label{fig:threestage}
  \vspace{-4mm}
\end{figure}

\paragraph{Limited Gain from Additional Stages.} \label{dis:multistage} A natural question arises: \textit{Will more stages continue to improve performance?}
Based on our experiments, the answer appears to be no. As illustrated in \Cref{fig:threestage}, model performance essentially plateaus at the beginning of a third stage, and while minor fluctuations are observed thereafter, the metrics do not exhibit new upward trends. Further training does not yield additional gains, likely due to diminishing learning signals.

\paragraph{Effect of Additional Steps at Inference Time. \label{dis:inference_flexibility}
}
Although DLM-One is optimized for single-step generation, we explore whether introducing additional steps at inference time can further enhance generation quality. Specifically, we implement a simple iterative scheme where the model alternates between re-noising and denoising its own output multiple times. As shown in \Cref{tab:multi_step_sampling}, increasing the number of steps from 1 to 4 consistently improves fidelity metrics (\textit{e.g.,} BLEU, ROUGE-L) at a modest cost to diversity (\textit{e.g.,} Div-4, self-BLEU). This demonstrates that even without explicit multi-step training, the number of sampling steps can serve as a practical lever to navigate the quality-diversity trade-off. However, since the model was not optimized for this regime, these results should not be interpreted as an upper bound. Training distilled generators specifically for few-step inference remains a promising direction inspired by recent vision models~\citep{yin2024improved, zhou2025few}.

\paragraph{Comparison with AR Models.}
A key advantage of our approach is that DLM-One maintains competitive generation quality compared to similar-sized AR LMs while offering a substantial improvement in inference speed. Our results align with prior work, which shows that teacher DLMs (\textit{e.g.,} DiffuSeq, 91M parameters) can already achieve on-par or superior performance to much larger, fine-tuned AR models like GPT-2 Large (774M parameters) across Seq2Seq tasks. By distilling the teacher into a one-step generator, DLM-One preserves this high fidelity while being orders of magnitude faster than both its multi-step teacher and the token-by-token sampling of AR models. We provide a detailed comparison of performance metrics, model sizes, and inference speeds in Appendix~\ref{app:ar_comparison}.

\begin{table}[!t]
\centering
\caption{\small
Performance of \ours{} under increased inference steps on the QQP dataset.}
\label{tab:multi_step_sampling}
\small
\begin{adjustbox}{max width=\linewidth}
\begin{tabular}{lccccccc}
\toprule
\textbf{Steps} & \textbf{BLEU}($\uparrow$) & \textbf{ROUGE-L}($\uparrow$) & \textbf{BERT}($\uparrow$) & \textbf{Dist-1}($\uparrow$) & \textbf{SelfBLEU}($\downarrow$) & \textbf{Div-4}($\uparrow$) \\
\midrule
1  & 0.1788 & 0.5265 & 0.7851 & 0.9671 & \textbf{0.3418} & \textbf{0.6256} \\
2  & 0.1800 & 0.5287 & 0.7895 & 0.9676 & 0.3455 & 0.6228 \\
4  & \textbf{0.1829} & \textbf{0.5329} & \textbf{0.7959} & \textbf{0.9693} & 0.3549 & 0.6095 \\
\bottomrule
\end{tabular}
\end{adjustbox}
\vspace{-4mm}
\end{table}

\section{Conclusion}
\label{sec:conclusion}

In this work, we propose a practical distillation framework for training continuous diffusion language models for one-step sequence generation (\ours{}), eliminating the need for iterative refinement during generation. Our method is broadly applicable to continuous DLMs and enables fast, one-step generation via score distillation from pretrained teacher models. To further stabilize training and improve student quality, we introduce a two-stage training scheme with adversarial regularization. Through detailed experiments on conditional text generation tasks, we demonstrate that \ours{} achieves competitive performance against the teacher DLMs while reducing sampling cost by up to $\mathord{\sim}500\times$. This redefines the Pareto frontier between computational efficiency and generation quality in continuous diffusion-based language modeling, and has profound implications for the future development of LLMs.

Nevertheless, our work opens up several promising directions for future investigation. First, while hyperparameters like the score distillation loss coefficient $\mu$ currently require per-task tuning, future work could explore more principled and adaptive training schemes. Second, we find that the trade-off between fidelity and diversity in \ours{} is a controllable aspect that can be adjusted based on the needs of downstream applications. We identify the extension of DLM-One to a few-step generator as a key avenue for future research, with the potential to improve the overall model performance in both fidelity and diversity.

\newpage

\small
\bibliography{references, zhougroup_ref, zhougroup_ref_1, proj_refs}
\bibliographystyle{plainnat}
\normalsize

\newpage
\appendix
\begin{center}
    {\textbf{\Large Appendix for \ours{}}}
\end{center}

\section{Broader Impacts}
\label{app:broader_impacts}
The high computational cost of large-scale language models poses challenges for accessibility, especially for users with limited resources. \ours{} addresses this by enabling one-step diffusion-based language generation, offering a significantly more efficient alternative to traditional iterative methods. By reducing the number of function evaluations required at inference time, \ours{} lowers energy consumption and makes diffusion language models more practical and sustainable for real-world deployment.

\section{Implementation Details \label{app:implementation_details}}

In this section, we provide detailed documentation of the implementation, including aspects not fully covered in the main text, for experiments on DiffuSeq. We outline the specific adaptations required for distilling these baselines into one-step sequence generators.

\subsection{DiffuSeq}
We adopt the official codebase of DiffuSeq\footnote{\url{https://github.com/Shark-NLP/DiffuSeq}} and all three released checkpoints to conduct our Seq2Seq experiments in \Cref{sec:experiments}. 

\subsubsection{Training Protocol}
For the training of our \ours{} student models, we set a fixed training budget of 50,000 steps for all datasets. We use AdamW~\citep{adamw} optimizer with $\beta_1=0.0$, $\beta_2=0.999$, and zero weight decay for both the student and the score estimator. The learning rate is fixed across tasks at $10^{-5}$. During Stage 1, we monitor the performance metrics on the validation set, such as BLEU, every 200 steps. Once the training is completed, we select the best-performing student checkpoint on the validation set as our new starting point for Stage 2. We provide a detailed table of distillation-related hyperparameters for both stages of each dataset in \Cref{tab:distill_hyperparams}.

\begin{table*}[!htbp]
\centering
\caption{Distillation-related hyperparameters used in Stage 1 and Stage 2 across different datasets.}
\label{tab:distill_hyperparams}
\small
\begin{tabular}{llccccccc}
\toprule
\textbf{Dataset} & \textbf{Stage} & $\mu$ & $[\tmin, \tmax]$ & $\tinit$ & $a^{sg}_{dsm}$, $b^{sg}_{adv}$ & $a^{g}_{sd}$, $b^{g}_{adv}$ & $\text{lr}_{\psi}$ & $\text{lr}_{\theta}$ \\
\midrule
\multirow{2}{*}{QQP} 
  & Stage 1 & 1.2 & $[0, 1976]$ & 1490 & 0.5, 0.5 & 0.5, 0.5 & 3e-5 & 1e-5 \\
  & Stage 2 & 0.5 & $[0, 1976]$ & 1490 & 0.5, 0.5 & 0.9, 0.1 & 1e-5 & 1e-5 \\
\midrule
\multirow{2}{*}{Q-T} 
  & Stage 1 & 1.2 & $[0, 1976]$ & 1490 & 0.5, 0.5 & 0.5, 0.5 & 1e-5 & 1e-5 \\
  & Stage 2 & 1.2 & $[0, 1976]$ & 1490 & 0.5, 0.5 & 0.5, 0.5 & 1e-5 & 1e-5 \\
\midrule
\multirow{1}{*}{Wiki} 
  & Stage 1 & 1.0 & $[0, 1976]$ & 1490 & 0.5, 0.5 & 0.5, 0.5 & 1e-5 & 1e-5 \\
\bottomrule
\end{tabular}
\end{table*}

\subsubsection{Conditioning}

During the pretraining of DiffuSeq models, the injection of conditions is achieved via concatenation, \textit{i.e.,} the condition sequence is directly concatenated with the data sequence as a whole before entering the network. 
However, the positions corresponding to the condition sequence do not participate in the diffusion forward process and are output as-is by the models.
To align with the teacher pretraining process, we adjust the output by the student model accordingly. Denote the condition embedding sequence as $e^{\text{cond}}$ and the initial noise for the student model $\theta$ as $z$. Let $\tilde{e}_{\theta,t}=G_\theta(e^{\text{cond}}, z)=e_\theta^{\text{cond}}\oplus e_\theta^{\text{data}}$. To inject the true condition, we modify the direct output by the student model (\textit{i.e.,} $\tilde{e}_{\theta,t}$) as $e_{\theta,t}=e^{\text{cond}}\oplus e_\theta^{\text{data}}$. The rationale behind this operation is that the teacher model has been trained on the true conditions from the real dataset only; using part of the generated sequence would introduce a discrepancy between teacher pretraining and distillation. Therefore, we replace the generated condition part, \textit{i.e.,} $e_\theta^{\text{cond}}$ with 
the true condition sequence $e^{\text{cond}}$. In our early experiments, we found that this adjustment helps stabilize training and prevent degeneration when used together with adversarial training.

\section{On the Fidelity-Diversity Trade-off}
\label{app:tradeoff}

The empirical results of the \ours{} model reflect a trade-off between generation fidelity and diversity. The proposed two-stage training process tends to prefer high-fidelity outputs over high-diversity outputs, which we argue is not necessarily a limitation for practical applications. In this section, we provide a detailed analysis of this trade-off and suggest how one can manage the trade-off in practice: 1) inference-time text augmentation to boost diversity, and 2) few-step generalization to increase \ours{}'s overall performance.

\subsection{Mitigating Diversity Loss with Text Augmentation}
\label{app:augmentation}

High fidelity is often preferable in practice, as many users want to call a model once and receive a high-quality, relevant answer. Our model is optimized for this single-call scenario, where higher fidelity metrics (\textit{e.g.,} BLEU, BERTScore) indicate stronger utility. While the resulting decrease in diversity might seem like a limitation, we demonstrate that it can be compensated for at inference time with a simple rule-based text augmentation.

Specifically, by randomly inserting a pad token (\texttt{[PAD]}) into the condition text with a given probability, we can directly boost the diversity of the generated sentences. \Cref{tab:tradeoff_full} presents the results of this experiment on PP, QG, and TS tasks. As the insertion probability increases, diversity metrics like Div-4 consistently rise across all tasks, accompanied by a predictable, modest decrease in fidelity scores. This shows that diversity in \ours{} is not a fixed limitation but rather a controllable parameter that can be tuned according to the needs of a specific application. We anticipate that with more advanced techniques, such as model-based augmentation, generation diversity could be further enhanced with even less impact on fidelity.

\begin{table}[h!]
\centering
\caption{The effect of random \texttt{[PAD]} token insertion on the fidelity-diversity trade-off across three tasks. As insertion probability increases, diversity (SelfBLEU and Div-4) consistently improves at the cost of fidelity.}
\label{tab:tradeoff_full}
\small
\begin{adjustbox}{max width=\linewidth}
\begin{tabular}{ll|c|cccc|c}
\toprule
\textbf{Task} & \textbf{Dataset} & \textbf{Ins. Prob.} & \textbf{BLEU}(↑) & \textbf{R-L}(↑) & \textbf{BERT}(↑) & \textbf{Dist-1}(↑) & \textbf{SelfB}(↓) / \textbf{Div-4}(↑) \\
\midrule
\multirow{3}{*}{PP} & \multirow{3}{*}{QQP} & 0.0 & 0.1788 & 0.5265 & 0.7851 & 0.9671 & 0.3418 / 0.6256 \\
& & 0.5 & 0.1746 & 0.5204 & 0.7798 & 0.9663 & 0.3224 / 0.6507 \\
& & 0.7 & 0.1712 & 0.5177 & 0.7771 & 0.9654 & 0.3134 / 0.6608 \\
\midrule
\multirow{3}{*}{QG} & \multirow{3}{*}{Q-T} & 0.0 & 0.1512 & 0.3257 & 0.5683 & 0.9053 & 0.6166 / 0.3798 \\
& & 0.5 & 0.1485 & 0.3175 & 0.5632 & 0.9065 & 0.5820 / 0.4167 \\
& & 0.7 & 0.1473 & 0.3144 & 0.5624 & 0.9064 & 0.5692 / 0.4294 \\
\midrule
\multirow{3}{*}{TS} & \multirow{3}{*}{Wiki} & 0.0 & 0.2927 & 0.5299 & 0.7565 & 0.8924 & 0.5456 / 0.4098 \\
& & 0.5 & 0.2769 & 0.5196 & 0.7486 & 0.8897 & 0.5015 / 0.4532 \\
& & 0.7 & 0.2715 & 0.5166 & 0.7464 & 0.8890 & 0.4866 / 0.4665 \\
\bottomrule
\end{tabular}
\end{adjustbox}
\end{table}

\subsection{Improving Diversity with Multi-Step Training}
\label{app:multi_step_training}

A more fundamental approach to improving both fidelity and diversity is to train the model for few-step generation. This would involve training the generator to perform denoising at different noise levels (\textit{i.e.,} step-aware training), significantly improving the model's ability to produce diverse results. Such strategies have proven highly effective for enhancing distilled student models in the vision domain~\citep{salimans2024multistep,zhou2025few} and represent a promising direction for future work on continuous DLMs.

\section{Comparison with Autoregressive (AR) Models}
\label{app:ar_comparison}

To evaluate the performance and efficiency of \ours{} against standard baselines, we compare it to its teacher model (DiffuSeq) and two fine-tuned autoregressive models (GPT-2 Base and GPT-2 Large). For inference speed, we report the average time in seconds over 100 runs on the text simplification task, with a maximum output length of 128 tokens for a fair comparison.

The results, shown in \Cref{tab:ar_comparison_full}, demonstrate a clear trade-off between model type, performance, and speed. Both DiffuSeq and our distilled DLM-One are competitive with or outperform the GPT-2 models on key fidelity metrics (BLEU, ROUGE-L, BERT), despite having significantly fewer parameters than GPT-2 Large. Most notably, DLM-One's single-step generation makes it by far the fastest model, achieving a speedup of approximately \textbf{27$\times$} over GPT-2 Base and \textbf{500$\times$} over its teacher, DiffuSeq.

\begin{table}[h!]
\centering
\caption{\textbf{Comparison of DLMs and AR models on the text simplification task.} DLM-One maintains competitive performance while being orders of magnitude faster. All results are reported using MBR-10 decoding for a fair comparison.}
\label{tab:ar_comparison_full}
\footnotesize
\setlength{\tabcolsep}{3pt}
\begin{adjustbox}{max width=\linewidth}
\begin{tabular}{l|cccccc|cc}
\toprule
\textbf{Model} & \textbf{BLEU}(↑) & \textbf{R-L}(↑) & \textbf{BERT}(↑) & \textbf{Dist-1}(↑) & \textbf{SelfB}(↓) & \textbf{Div-4}(↑) & \textbf{\# Params} & \thead{\textbf{Avg. Inf.}\\\textbf{Time (s)}} \\
\midrule
GPT-2 Base FT & 0.3083 & 0.5461 & 0.8021 & 0.9439 & 0.5444 & 0.6047 & 117M & 0.82 \\
GPT-2 Large FT & 0.2693 & 0.5111 & 0.7882 & \textbf{0.9464} & 0.6042 & 0.5876 & 774M & 2.34 \\
\midrule
DiffuSeq (Teacher) & 0.3622 & \textbf{0.5849} & \textbf{0.8126} & 0.9264 & \textbf{0.4642} & \textbf{0.6604} & \textbf{91M} & 14.94 \\
DLM-One (Student) & \textbf{0.3630} & 0.5839 & 0.8084 & 0.9068 & 0.5456 & 0.4098 & \textbf{91M} & \textbf{0.03} \\
\bottomrule
\end{tabular}
\end{adjustbox}
\end{table}

\section{Pseudo Code}
See \Cref{algo:main}.

\begin{algorithm}[htbp]
\caption{
DLM-One Adversarial Score Distillation
}\label{algo:main}
\begin{algorithmic}
\small
\STATE \textbf{Input:} Pre-trained teacher DLM $\phi$, student model $\theta$, score estimator $\psi$, embedding layer $E$, score distillation loss coefficient $\mu$, real dataset $\cD_{X,C}$, time range $[\tmin, \tmax]$, diffusion weight function $\lambda(t)$, loss term coefficients $\sgdsm{a},\sgadv{b},\gsid{a},\gadv{b}$.

\STATE \textbf{Initialization} $\theta \gets \phi, \psi \gets \phi$ 
\REPEAT
\STATE Sample $c^{\text{fake}}\sim\cD_{*,Y},\,(x^{\text{real}},c^{\text{real}}) \sim \cD_{X,C},\,t\in [\tmin, \tmax ]$ 
\STATE Sample $z\sim\mathcal N(0,\mathbf I)$, let $e^{\text{fake}}=G_\theta(c^{\text{fake}},z)$ and $e^{\text{real}}=E(x^{\text{real}})$
\STATE Sample noises $\eps^{\text{fake}},\eps^{\text{real}}\sim\whitenoise$
\STATE $e_t^{\text{fake}}\gets\alpha_te^{\text{fake}}+\sigma_t\eps^{\text{fake}},e_t^{\text{real}}\gets\alpha_te^{\text{real}}+\sigma_t\eps^{\text{real}}$
\STATE Compute $\hat{\cL}_{\text{dsm}}$ according to Eq.\,\ref{eq:dsm} and $\hat{\cL}_{\text{adv}}^{sg}$ according to Eq.\,\ref{eq:sg_adv}
\STATE Update $\psi$ via SGD on the combined loss $\sgdsm{a}\hat{\cL}_{\text{dsm}}+\sgadv{b}\sgadv{\hat{\cL}}$
\STATE Sample $c^{\text{fake}}\sim\cD_{*,C},\,t\in [\tmin, \tmax ]$
\STATE Sample $z\sim\mathcal N(0,\mathbf I)$, let $e^{\text{fake}}=G_\theta(c^{\text{fake}},z)$ %
\STATE Sample noises $\eps^{\text{fake}}\sim\whitenoise$
\STATE $e_t^{\text{fake}}\gets\alpha_te^{\text{fake}}+\sigma_t\eps^{\text{fake}}$
\STATE Compute $\hat{\cL}_{\text{sd}}$ according to Eq.\,\ref{eq:sid} and $\hat{\cL}_{\text{adv}}^{g}$ according to Eq.\,\ref{eq:g_adv}
\STATE Update $\theta$ via SGD on the combined loss $\gsid{a}\hat{\cL}_{\text{sd}}+\gadv{b}\gadv{\hat{\cL}}$
\UNTIL{the maximum number of training steps is reached}
\STATE \textbf{Output:} $\theta$
\normalsize
\end{algorithmic} 
\end{algorithm}

\section{Case Study: TESS Diffusion}
\label{app:tess_details}

To verify the robustness of \ours{} beyond standard Euclidean embedding-space models, we conducted an extension study on \textbf{TESS}~\citep{mahabadi2024tess}, which is a continuous diffusion model that performs text-to-text generation on a \textit{k-logit simplex space} rather than a typical word embedding space. This presents a unique challenge for one-step distillation as the model must navigate a probability manifold.

\paragraph{Methodology and Results.} We applied the score distillation principle to the simplex representations predicted by the TESS backbone. Our results on the paraphrase task demonstrate that \ours{} successfully distills the 1000-step TESS teacher into a 1-step student. As shown in Table~\ref{tab:tess_results}, while the teacher achieves a high-fidelity BLEU score of 0.3037, the one-step student preserves a competitive score of 0.2514 while reducing the number of function evaluations (NFEs) from 1000 to 1.

\begin{table}[h]
\centering
\caption{\textbf{Quantitative results for the TESS case study on the QQP dataset.} \ours{} achieves a $1000\times$ speedup on the paraphrase task while preserving significant generative quality.}
\label{tab:tess_results}
\small
\begin{adjustbox}{max width=\linewidth}
\begin{tabular}{lccccc}
\toprule
\textbf{Model} & \textbf{BLEU}($\uparrow$) & \textbf{R-L}($\uparrow$) & \textbf{BERT}($\uparrow$) & \textbf{Dist-1}($\uparrow$) & \textbf{NFEs}($\downarrow$) \\
\midrule
TESS (Teacher) & 0.3037 & 0.6182 & 0.9496 & 0.9860 & 1000 \\
\ours{} (Student) & 0.2514 & 0.5863 & 0.9380 & 0.9350 & \textbf{1} \\
\bottomrule
\end{tabular}
\end{adjustbox}
\end{table}

\paragraph{Scaling Potential.} The success of this adaptation is significant as TESS serves as the foundational framework for \textbf{TESS-2}~\citep{tae2025tess2}, a large-scale generalist diffusion model that has demonstrated performance parity with state-of-the-art autoregressive models. These results provide a proof-of-concept that \ours{} can serve as an orthogonal acceleration layer for the next generation of high-performing diffusion generalists.

\section{Additional Results}
Due to the page limit of the main text, we defer supplementary experimental results to this section.

\subsection{Generated Samples for Seq2Seq Tasks}
We present generation results on 5 random examples each from the PP, QG, and TS tasks in \Cref{tab:pp_examples,tab:qg_examples,tab:ts_examples}.

\subsection{\ours{} with MBR Decoding}
To directly compare with the results reported in \citet{gong2023diffuseq}, we evaluate our student models using the MBR decoding strategy with a total of 10 generated candidates (denoted as MBR-10). As shown in \Cref{tab:mbr10}, our distilled models demonstrate comparable performance to their respective teachers across all three datasets (QQP, QG, Wiki). In particular, the student model on the Wiki dataset nearly matches the teacher in all quality metrics (BLEU, ROUGE-L, BERTScore), suggesting that the \ours{} model can retain strong performance even when evaluated using multiple samples. However, we also observe a decrease in diversity metrics, especially on QG, which indicates that MBR may favor models with higher inter-sentence diversity.

\section{LLM Usage Statement}
We utilized large language models (\textit{e.g.,} ChatGPT) to assist with proofreading, grammatical corrections, and polishing the text of this manuscript. No new scientific results or text contents were generated by LLMs.

\begin{table*}[t]
\centering
\caption{\textbf{MBR-10 evaluation results across Seq2Seq tasks.} Arrows indicate preferred directions: $\uparrow$ higher is better, $\downarrow$ lower is better.}
\label{tab:mbr10}
\small
\begin{adjustbox}{max width=\linewidth}
\begin{tabular}{ll|lccccc}
\toprule
\textbf{Task} & \textbf{Dataset} & \textbf{Model} & \textbf{BLEU}(↑) & \textbf{R-L}(↑) & \textbf{BERT}(↑) & \textbf{Dist-1}(↑) & \textbf{SelfB}(↓) / \textbf{Div-4}(↑) \\
\midrule
\multirow{2}{*}{PP} & \multirow{2}{*}{QQP} 
& DiffuSeq & 0.2413 & 0.5880 & 0.8365 & 0.9807 & 0.2732 / 0.8641 \\
& & \ours{} & 0.2213 & 0.5741 & 0.8297 & 0.9773 & 0.3418 / 0.6256 \\
\midrule
\multirow{2}{*}{QG} & \multirow{2}{*}{Q-T} 
& DiffuSeq & 0.1731 & 0.3665 & 0.6123 & 0.9056 & 0.2789 / 0.8103 \\
& & \ours{} & 0.1522 & 0.3280 & 0.5708 & 0.9026 & 0.6167 / 0.3798 \\
\midrule
\multirow{2}{*}{TS} & \multirow{2}{*}{Wiki} 
& DiffuSeq & 0.3622 & 0.5849 & 0.8126 & 0.9264 & 0.4642 / 0.6604 \\
& & \ours{} & 0.3630 & 0.5839 & 0.8084 & 0.9068 & 0.5456 / 0.4098 \\
\bottomrule
\end{tabular}
\end{adjustbox}
\end{table*}

\begin{table*}[t]
\centering
\caption{\textbf{Examples from the Paraphrase (PP) task.} Each example consists of a source sentence, a reference sentence, and outputs generated by DiffuSeq (Teacher) and DLM-One (Student).}
\label{tab:pp_examples}
\small
\begin{tabular}{p{0.20\linewidth}p{0.20\linewidth}p{0.20\linewidth}p{0.20\linewidth}}
\toprule
\multirow{2}{*}{\textbf{Source}} & \multirow{2}{*}{\textbf{Reference}} & \multicolumn{2}{c}{\textbf{Recover}} \\
\cmidrule(lr){3-4}
& & \textbf{DiffuSeq} & \textbf{DLM-One} \\
\midrule
how can i be a good geologist? & what should i do to be a great geologist? & how do i really be a good geologist? & how can i become a good geologist? \\ \midrule
which are the best engineering fields? & what is the best field of engineering? & which are the best engineering field? & what are the best engineering fields? \\ \midrule
how do i become an attractive girl? & how do you become pretty / attractive? & how can one become a girl? & how can i become an attractive girl quickly? \\ \midrule
how does a long distance relationship work? & do long distance relationships work? & does long distance relationship work? & how do i have a long distance relationship? \\ \midrule
what are some interesting things to do when bored? & what should i do if i'm badly bored? & what should you do when you bored? & what are the best thing to do when bored? \\
\bottomrule
\end{tabular}
\end{table*}

\begin{table*}[t]
\centering
\caption{\textbf{Examples from the Question Generation (QG) task.} Each example consists of a source sentence, a reference sentence, and outputs generated by DiffuSeq (Teacher) and DLM-One (Student).}
\label{tab:qg_examples}
\small
\begin{tabular}{p{0.20\linewidth}p{0.20\linewidth}p{0.20\linewidth}p{0.20\linewidth}}
\toprule
\multirow{2}{*}{\textbf{Source}} & \multirow{2}{*}{\textbf{Reference}} & \multicolumn{2}{c}{\textbf{Recover}} \\
\cmidrule(lr){3-4}
& & \textbf{DiffuSeq} & \textbf{DLM-One} \\
\midrule
a gaggle is a group of geese. & what is a group of geese called & what kind of birds would you a group geese geese & what is a group of geese called? \\ \midrule
the ten - mineral mohs scale of relative hardness, based on what scratches what. & what is measured by moh's scale? & in minerology what does the mohs scale measure & in minerology what does the mohs scale measure \\ \midrule
if you mix red and green lights they do not magically change into yellow light. & what colour do you get when you mix blue and yellow together? & when you mix equal amounts of blue and yellow color do what color? & when you mix equal amounts of blue and yellow yellow, what color do you get? \\ \midrule
capable of sustained hovering, the hummingbird has the ability to fly deliberately backwards & which is the only musical bird that can fly backwards & what is the only bird that can can fly backwards & what is the only bird that can fly backwards \\ \midrule
alexander graham bell in 1876, at the age of 29, alexander graham bell invented his telephone. & what did alexander graham bell invent & the telephone was invented in which year & the telephone was invented in which year \\
\bottomrule
\end{tabular}
\end{table*}

\begin{table*}[t]
\centering
\caption{\textbf{Examples from the Text Simplification (TS) task.} Each example consists of a source sentence, a reference sentence, and outputs generated by DiffuSeq (Teacher) and DLM-One (Student).}
\label{tab:ts_examples}
\small
\begin{tabular}{p{0.20\linewidth}p{0.20\linewidth}p{0.20\linewidth}p{0.20\linewidth}}
\toprule
\multirow{2}{*}{\textbf{Source}} & \multirow{2}{*}{\textbf{Reference}} & \multicolumn{2}{c}{\textbf{Recover}} \\
\cmidrule(lr){3-4}
& & \textbf{DiffuSeq} & \textbf{DLM-One} \\
\midrule
she was also the leader of the party between 1993 and 1995. & she was also the leader of the party between 1993 and 1995. & she was the leader of the party from 1995 to 1993. & she was the leader between 1993 and 1995. \\ \midrule
thiel - sur - acolin is a commune in the allier department in auvergne - rhone - alpes in central france. & thiel - sur - acolin is a commune. & thiel - sur - acolin is a commune. & thiel - sur - acolin is a commune. \\ \midrule
vetlanda municipality ( " vetlanda kommun " ) is a municipality in jonkoping county, in southern sweden where the town of vetlanda is the seat. & vetlanda municipality is a municipality in jonkoping county in southern sweden. & vetlanda municipality is a municipality in jonkoping county in southern sweden. & vetlanda municipality is a municipality in jonkoping county. \\ \midrule
beaufort is located in north carolina's " inner banks " region. & beaufort is in north carolina's inner banks region. & beaufort is in north carolina's " inner banks " region. & beaufort is located in " inner banks " region.  \\ \midrule
weaver was born in pittsburgh, pennsylvania, on january 19, 1926, the son of elsa w. ( nee stringaro ) weaver and john carson weaver. & weaver was born on january 19, 1926 in pittsburgh, pennsylvania. & weaver was born in pittsburgh, pennsylvania, on january 19, 1926. & weaver was born in pittsburgh, pennsylvania. \\
\bottomrule
\end{tabular}
\end{table*}

\end{document}